\documentclass[conference]{IEEEtran}
\IEEEoverridecommandlockouts
\usepackage{cite}
\usepackage{amsmath,amssymb,amsfonts}
\usepackage{graphicx}
\usepackage{textcomp}
\usepackage{xcolor}
\usepackage[ruled,linesnumbered]{algorithm2e}
\usepackage{multirow}
\usepackage{booktabs}

\def\BibTeX{{\rm B\kern-.05em{\sc i\kern-.025em b}\kern-.08em
    T\kern-.1667em\lower.7ex\hbox{E}\kern-.125emX}}
\begin{document}

\title{Multi-Relation Graph-Kernel Strengthen Network for Graph-Level Clustering
}

\author{

\IEEEauthorblockN{
Renda Han \IEEEauthorrefmark{1}\textsuperscript{1}, 
Guangzhen Yao \IEEEauthorrefmark{2}\IEEEauthorrefmark{5}\textsuperscript{1},
Mingyuan Li \IEEEauthorrefmark{4}\textsuperscript{1},
Wenxin Zhang \IEEEauthorrefmark{9}\IEEEauthorrefmark{5}\textsuperscript{1},
Yu Li \IEEEauthorrefmark{6}\textsuperscript{1},\\
Wen Xin \IEEEauthorrefmark{1}\textsuperscript{2},
Huajie Lei \IEEEauthorrefmark{1}\textsuperscript{3}, 
Mengfei Li \IEEEauthorrefmark{1}\IEEEauthorrefmark{5}\textsuperscript{4},
Zeyu Zhang \IEEEauthorrefmark{7}\textsuperscript{1},
Chengze Du \IEEEauthorrefmark{2}\textsuperscript{1},
and Yahe Tian\IEEEauthorrefmark{8}\textsuperscript{1}} 

\IEEEauthorblockA{\IEEEauthorrefmark{1} Hainan University, Haikou, China\\ 
\textsuperscript{1}hanrenda@hainanu.edu.cn, \textsuperscript{2}614751955@qq.com,
\textsuperscript{3}leihuajie@hainanu.edu.cn, 
\textsuperscript{4}20233005226@hainanu.edu.cn,}

\IEEEauthorblockA{\IEEEauthorrefmark{2}Northeast Normal University, Changchun, China
\textsuperscript{1}yaoguangchen@nenu.edu.cn}

\IEEEauthorblockA{\IEEEauthorrefmark{4}Hebei University of Technology, Tianjin, China
\textsuperscript{1}1372349386@qq.com}

\IEEEauthorblockA{\IEEEauthorrefmark{9}University of Chinese Academy of Science, Beijing, China
\textsuperscript{1}zwxzhang12@163.com}

\IEEEauthorblockA{\IEEEauthorrefmark{6}Hubei University, Wuhan, China \textsuperscript{1}746433542@qq.com}

\IEEEauthorblockA{\IEEEauthorrefmark{7}The Australian National University, Canberra, Australia \textsuperscript{1}steve.zeyu.zhang@outlook.com}

\IEEEauthorblockA{\IEEEauthorrefmark{2}Beijing University of Posts and Telecommunications, Beijing, China
\textsuperscript{1}ducz0338@bupt.edu.cn}

\IEEEauthorblockA{\IEEEauthorrefmark{8}Southeast University, Nanjing, China
\textsuperscript{1}2445275333@qq.com}

\IEEEauthorblockA{\IEEEauthorrefmark{5}Corresponding Author}}

%\author{\IEEEauthorblockN{Anonymous Authors}}

\maketitle

\begin{abstract}
Graph-level clustering is a fundamental task of data mining, aiming at dividing unlabeled graphs into distinct groups. However, existing deep methods that are limited by pooling have difficulty extracting diverse and complex graph structure features, while traditional graph kernel methods rely on exhaustive substructure search, unable to adaptively handle multi-relational data. This limitation hampers producing robust and representative graph-level embeddings. To address this issue, we propose a novel \underline{M}ulti-Relation \underline{G}raph-Kernel \underline{S}trengthen \underline{N}etwork for Graph-Level Clustering (MGSN), which integrates multi-relation modeling with graph kernel techniques to fully leverage their respective advantages. Specifically, MGSN constructs multi-relation graphs to capture diverse semantic relationships between nodes and graphs, which employ graph kernel methods to extract graph similarity features, enriching the representation space. Moreover, a relation-aware representation refinement strategy is designed, which adaptively aligns multi-relation information across views while enhancing graph-level features through a progressive fusion process. Extensive experiments on multiple benchmark datasets demonstrate the superiority of MGSN over state-of-the-art methods. The results highlight its ability to leverage multi-relation structures and graph kernel features, establishing a new paradigm for robust graph-level clustering.
\end{abstract}


\section{Introduction}
Graph-level clustering \cite{li2025multi} plays a pivotal role in understanding complex, high-dimensional data by identifying patterns \cite{liang2024mgksite} and relationships across multi-graphs. This approach is particularly valuable in domains such as bioinformatics \cite{Zhang2018GraphBasedAnalysis}, recommender systems \cite{Ding2018MultiRelationalGraphEmbedding}, and social media analysis \cite{Kumar2017GraphBasedCommunityDetection, cai2024lgfgad}, where the latent structure of graph-level holds key insights. By focusing on graph-level features, this method more effectively organizes and analyzes complex problems, informing decision-making in areas such as disease prediction \cite{zhang2023graph}, user behavior \cite{jiang2022graph}, and content categorization \cite{li2024comprehensive}. Moreover, it is more effective in uncovering latent patterns that are not readily identifiable.

The existing graph-level clustering methods are mainly divided into two categories by learning mode: traditional graph kernel methods based on spectral clustering and deep methods based on representation learning.
Traditional graph kernel methods \cite{kashima2003marginalized, borgwardt2005shortest, shervashidze2009efficient, shervashidze2011weisfeiler, ramon2003expressivity} measure similarity between graphs using predefined structural features, such as random walks or shortest paths. These methods excel in capturing the structural features of smaller graphs and are effective for graphs with well-defined, simple relationships. However, they face significant challenges when applied to large, heterogeneous graphs, especially those with multiple relational types, due to scalability limitations. In contrast, deep learning-based methods have emerged as powerful tools for learning graph representations. Graph Convolutional Networks (GCNs) \cite{liang2024hawkes, liu2024learn, liang2024simple} automatically learn from raw graph data, capturing complex, non-linear relationships between nodes while maintaining scalability, which makes them suitable for a wide range of large-scale applications. These graph-level clustering methods \cite{sun2019infograph, you2020graphcl, cai2024deep, ju2023glcc} enhance the robustness of graph representations by leveraging contrastive learning, multi-view clustering, and automated data augmentation techniques, making them well-suited for complex and diverse datasets. They capture both global semantics and local structures within graphs, improving their ability to represent intricate graph relationships.
% Additionally, by combining contrastive learning with clustering, they optimize performance in tasks like graph classification and clustering, particularly when dealing with graphs that exhibit diverse relational patterns. 
These methods aggregate node-level features into a global graph-level representation, enabling them to scale more effectively and adapt to more complex scenes.

However, methods in both categories encounter significant scalability challenges. Traditional graph kernel approaches, which depend on exhaustive substructure searches to assess graph similarity, often fall short in capturing latent dependencies among relational types. These dependencies are crucial for accurately constructing the graph-level representation. By neglecting the intricate multi-relational dynamics within graphs, traditional methods fail to encapsulate their underlying semantics, thereby limiting their effectiveness in processing multi-relational graph data.
On the other hand, while deep learning methods have overcome many limitations of traditional approaches by automatically learning important features and relationships in graph structures, they are still constrained by pooling operations. These operations, which reduce dimensionality and aggregate node features, often result in the loss of crucial structural details, hindering the model’s ability to capture complex relational patterns within graphs. 

To solve this issue, inspired by multi-relation clustering \cite{shen2024balanced}, we combine deep learning-based methods with graph kernel methods to offer a Multi-relation Graph-kernel Strengthen Network for graph-level clustering (MGSN) to overcome the individual limitations of each approach. Specifically, the model generates graph representations using a GCN, which serves as input for spectral clustering to construct multi-relational graphs. By incorporating various relational types, it captures the semantic relationships between nodes and graphs, effectively identifying the complex interactions within the graph structure. This approach allows for a more accurate representation of the multi-relational features inherent in the graph.
Moreover, a relation-aware representation refinement strategy is proposed, which optimizes graph-level features by progressively fusing information from multiple relations to adaptively establish inherent connections, ensuring that the refined embeddings capture complex relational patterns and improve the overall quality of the graph-level embeddings. 
In summary, our contributions are as follows:
%The fusion process is designed to enhance the representation incrementally, allowing the model to better integrate diverse relational information and ultimately improve the performance of graph-level clustering.

% \cite{cai2021graphclustering, zhang2022kernelmethods}.

\begin{itemize}  
\item A Multi-relation Graph-kernel Strengthen Network (MGSN) is proposed, which integrates multi-relation modeling and graph kernel techniques to capture diverse semantic relationships and enrich the representation space with graph similarity features, which ensures the consistency and discriminability of the learned representations.
\item A relation-aware representation refinement strategy is designed, adaptively aligning multi-relation information across views and progressively fusing complementary features through a kernel-guided alignment mechanism, preserving structure nuances. 
\item Extensive experiments on multiple benchmark datasets demonstrate that MGSN achieves superior performance compared to state-of-the-art methods, highlighting its capability to leverage multi-relation structures and dynamic graph kernel features effectively.  
\end{itemize}

%\section{Materials and Methods}
%Materials and Methods should be described with sufficient details to allow others to replicate and build on published results. Please note that the publication of your manuscript implies that you must make all materials, data, computer code, and protocols associated with the publication available to readers. Please disclose at the submission stage any restrictions on the availability of materials or information. New methods and protocols should be described in detail, while well-established methods can be briefly described and appropriately cited.

%Research manuscripts reporting large datasets that are deposited in a publicly available database should specify where the data have been deposited and provide the relevant accession numbers. If the accession numbers have not yet been obtained at the time of submission, please state that they will be provided during the review. They must be provided before publication.

\section{Related Work}
\subsection{Graph Kernels Method}
Graph kernels have been fundamental in graph-level analysis, providing a mathematically rigorous measure of graph similarity. Early methods, such as the Random Walk Kernel \cite{kashima2003marginalized} and Shortest Path Kernel \cite{borgwardt2005shortest}, focused on comparing substructures like walks and paths but suffered from high computational complexity and sensitivity to graph size. To address these issues, Graphlet Kernels \cite{shervashidze2011weisfeiler} and Subtree Kernels \cite{shervashidze2010subtree} shifted the focus to local patterns, improving efficiency while maintaining expressiveness. However, scalability and effectiveness in handling attributed graphs remained challenges.  The Weisfeiler-Lehman (WL) Kernel \cite{shervashidze2011weisfeiler} introduced an iterative tree aggregation process, significantly enhancing efficiency and adaptability to attributed graphs. Despite its success, WL Kernels rely on fixed-feature engineering and struggle to capture higher-order structures. To overcome these limitations, we propose integrating graph kernels with deep learning, leveraging graph kernels’ ability to capture structural similarities while enhancing representation learning. 
% This integration enables more expressive and adaptive feature extraction, improving multi-relational feature learning and refining clustering boundaries.

\subsection{Deep Graph-level Clustering}
With the rise of GCN, deep clustering methods based on representation learning have become crucial for graph structure analysis. Early approaches like InfoGraph \cite{sun2019infograph} captured local and global semantics by maximizing mutual information but struggled with complex graph relationships. GraphCL \cite{you2020graphcl} introduced multi-view contrastive learning to enhance relational modeling but faced optimization challenges in high-dimensional spaces. To address this, JOAO \cite{you2021graph} automated data augmentation selection, improving contrastive learning efficiency, though its reliance on augmentation strategies limited generalizability. GWF \cite{Xu2022} further advanced graph-level feature fusion, enhancing representation expressiveness. GLCC \cite{ju2023glcc} unified contrastive learning and clustering, refining graph-level representations for diverse relational patterns. Additionally, DCGLC \cite{cai2024dual} introduced a contrastive mechanism to align clustering information, capturing comprehensive cluster structures.
However, clustering graphs with highly diverse structural characteristics remains challenging, as structural variations can greatly affect performance.

\section{Method}
\begin{figure*}[!htbp]
    \centering
    \includegraphics[width=1\linewidth]{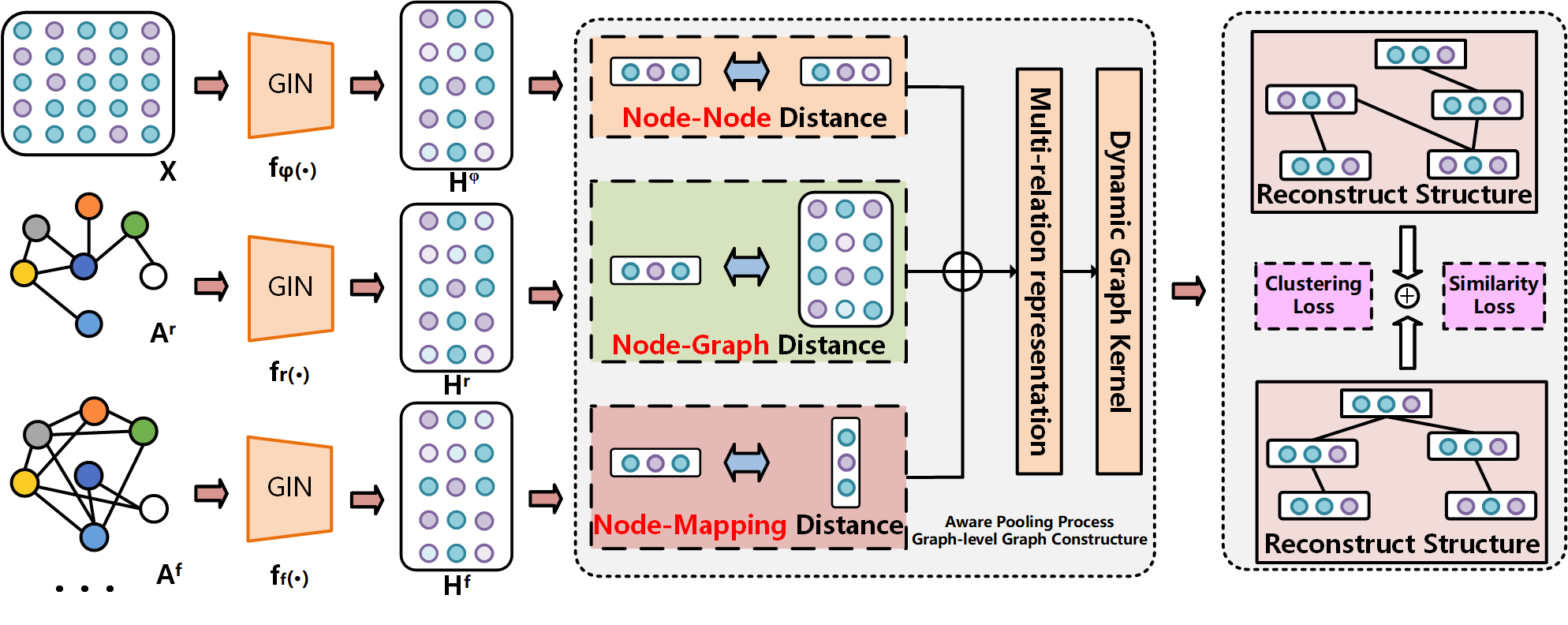}
    \caption{The overall architecture of MGSN. The method effectively utilizes multi-relational features to construct graph-level graphs and collaboratively guides the generation of high-quality representations with discriminability.}
    \label{fig:framework}
    \vspace{-10pt}
\end{figure*}

In this section, we provide a detailed description of MGSN, which integrates multi-relation graph construction, representation learning, and graph kernel strengthening. This combination facilitates the learning of highly discriminative graph-level representations, ultimately improving clustering performance. MGSN is explained in five main parts: multi-eelation graph construction, graph presentation encoding, aware pooling process, graph-level graph construction, and optimization objective, with the corresponding pseudo-code presented in Algorithm 1. The overall architecture is shown in Fig.~\ref{fig:framework}.

\begin{algorithm}[t]
\caption{Training process of MGSN}
\LinesNumbered 

\KwIn {$\mathbf{X}$, Graph set: $\mathcal{G}$, Number of clusters: $C$, Batch size $N_b$, Train epoch: $MaxIter$.}
\KwOut{Clustering results $\mathbf{R}$. }
Initialize  $\mathbf{W}^{(l)}_{gin}$ randomly\;
%\State Generate structure embedding $\mathbf{X}_s$

Construct multi-relation graph structure $\mathbf{A}^r$, fusion graph $\mathbf{A}^f$ by Eq. (1)-(3)\;
\For{iter = 1 to $MaxIter$}{
Generate node-level final embedding $\mathbf{\widetilde{H}}^\phi$,  $\mathbf{\widetilde{H}}^r$, $\mathbf{\widetilde{H}}^f$ by Eq. (4)\;
Pool these embedding to generate graph-level representation $\mathbf{Z}^\phi$, $\mathbf{Z}^r$,$\mathbf{Z}^f$ by Eq. (5)-(7)\;
Utilize dynamic graph kernel to construct graph-level graphs\;
Calculate self cluster loss $\mathcal{L}_{clu}$ by Eq. (11)\;
Calculate contrastive loss $\mathcal{L}_{con}$ by Eq. (12)\;
Calculate graph similarity loss $\mathcal{L}_{sim}$ by Eq. (13)\;
Backpropagate and update parameters\;
}
Calculate the clustering results by final embedding by K-Means\;
\KwRet{ $\mathbf{R}$ }
\end{algorithm}

\subsection{Multi-Relation Graph Construction}
To generate a multi-relational graph, we define multiple types of relations among nodes and edges that reflect specific structural or attribute information. Specifically, the original graph is \( G = (V, E) \), where \( V \) represents the set of nodes and \( E \) represents the set of edges. For each relation \( r \) in the predefined set of relations \( \mathcal{R} \), a relation graph \( G^r = (V, E^r) \) is constructed, where \( E^r \) consists of edges that satisfy the criteria of relation \( r \). \( G^\phi = (V, E^\phi) \) is origin graph. To facilitate the description of the following method, we make the following definition: $\mathbf{X}$ is the origin attribute matrix for each graph from the origin graph. $\mathbf{A}^r$ is the adjacent matrix with different relation $r$ for each graph. $\mathbf{D}^r$ is the degree matrix with different relation $r$ for each graph. Then, we construct multi-relation graphs through attribute-based similarity and grouping mechanisms. $a^r_{uv} \in \mathbf{A}^r$ can be defined as
\begin{equation}
\mathbf{a}^{r}_{uv} = \frac{\mathbf{x}_u \cdot \mathbf{x}_v}{\|\mathbf{x}_u\| \|\mathbf{x}_v\|},
\end{equation}
and edge relations are defined  \( \mathbf{e}_{uv} \) and \( \mathbf{e}_{u'v'} \):
\begin{equation}
\text{Dist}(\mathbf{e}_{uv}, \mathbf{e}_{u'v'}) = \|\mathbf{e}_{uv} - \mathbf{e}_{u'v'}\|_2,
\end{equation}
where Dist$(\cdot)$ is the function that calculates the distance between edge features.
The structure information from different relational graphs can be fused by combining their adjacency matrices in a weighted manner. For each relation \( r \), we have the corresponding adjacency matrix \( \mathbf{A}^r \). The final adjacency matrix can be obtained by weighted averaging as

\begin{equation}
\mathbf{A}^{f} = \sum_{r \in \mathcal{R}} \alpha_r \cdot \mathbf{A}^r
\end{equation}
where \( \alpha_r \) is the learnable weight assigned to each relation, which combines structure information from different relations into a unified structure representation.

\subsection{Graph Presentation Encoding}
To obtain more effective graph representations through multi-relational graphs, we leverage the Graph Isomorphism Network (GIN) for feature encoding. Given the set of multi-relational adjacency matrices \( \{\mathbf{A}^r \mid r \in \mathcal{R}\} \) and a shared node attribute matrix \( \mathbf{X} \), the GIN information transmission process can be described as:
$
\mathbf{H}^{(l+1)} = \sigma\left( \mathbf{W} \left( \mathbf{X} + \widehat{\mathbf{A}} \mathbf{H}^{(l)} \right) + \mathbf{b} \right),
$
where, \( \mathbf{H}^{(l)} \) is the embedding matrix of node features for relation of $l$-th layer, $\mathbf{H}^{(0)}$ is $\mathbf{X}$,  \( \mathbf{\widehat{A}} \) is the normalized adjacency matrix \( \mathbf{W} \) is the learnable weight matrix, \( \mathbf{b}^r \) is the bias matrix, \( \sigma \) is a non-linear activation function (e.g., ReLU). Here, $\mathbf{A}$ is only used as a representative symbol, and its actual value is $\mathbf{A}^\phi$, $\mathbf{A}^r$ and $\mathbf{A}^f$. In summary, three node-level representations can be generated, including original representation $\mathbf{\widetilde{H}}^{\phi}$, multi-relation representation $\mathbf{\widetilde{H}}^{r}$, and fusion representation $\mathbf{\widetilde{H}}^{f}$.

\subsection{Aware Pooling Process}
To capture both node-level and graph-level similarities, we propose an advanced pooling mechanism that extends beyond conventional methods by considering both pairwise node similarities and the interactions between individual nodes and the global graph structure. This approach effectively captures each node’s relative importance and structural affinities within the graph. 
% As a result, it enhances graph-level representation, improving the model’s ability to identify complex structural patterns and node relationships.
Specifically, we compute an importance score \( s_v \) for each node \( v \) via $
s_v = \text{softmax} \left( \mathbf{q}^\top \mathbf{h}_v^r \right)
$, which reflects its significance within the graph. \( \mathbf{q} \) is a shared learnable query vector. Next, we calculate the structure similarity between nodes by measuring the difference between node embeddings \( \mathbf{h}_u \) and \( \mathbf{h}_v \), which quantifies the similarity between node pairs. The structure similarity matrix \( \mathbf{S} \), where each element \( s_{uv} \in \mathbf{S} \) denotes the similarity between nodes \( u \) and \( v \) as
\begin{equation}
s_{uv} = \text{exp}\left( -\text{Dist}(e_{u}, e_{v}) \frac{\|\mathbf{h}_u - \mathbf{h}_v\|_2^2}{\tau} \right),
\end{equation}

where \( \tau \) is a temperature hyperparameter. To quantify the similarity between individual nodes and the graph as a whole, we compute the relationship between the graph-level representation \( \mathbf{\bar{z}}_i \) and each node embedding \( \mathbf{h}_v \) within the graph. This similarity provides a quantitative assessment of the alignment between node embeddings and the aggregated graph-level representation. The similarity can be expressed as
\begin{equation}
\text{Sim}(v, G) = \frac{\mathbf{h}_v^T \mathbf{\bar{z}}_i}{\|\mathbf{h}_v\|_2 \|\mathbf{\bar{z}}_i\|_2} \text{ s.t. } v \in G_i,
\end{equation}
where, \( \mathbf{\bar{z}}_{i} \) is the graph-level average pooling representation for $i$-th graph. The node-graph similarity measures how important and contributive a node is to the overall graph structure.

By integrating node scores, structural similarities, and node-to-graph similarities, we employ a weighted aggregation strategy to derive the graph-level representation \( \mathbf{z}_u \) for the \( u \)-th graph. Specifically, node embeddings are aggregated according to their importance, pairwise structural relationships, and overall relevance to the graph. $\mathbf{z}_u$ is computed as

\begin{equation}
\mathbf{z}_u = \sum_{v \in V} s_v \cdot \mathbf{h}_v + \cdot \sum_{v \in V} \sum_{u \in V} s_{uv} \cdot \mathbf{h}_u + \cdot \sum_{v \in V} \text{Sim}(v, G) \cdot \mathbf{h}_v.   
\end{equation}

The first term represents the importance-weighted node embeddings, the second term corresponds to the structure-weighted node embeddings, and the third term captures the node-graph similarity-weighted embeddings. In this way, we derive three graph-level representations: the original graph-level representation \( \mathbf{Z^\phi} \); the graph-level representation  \( \mathbf{Z_G^r} \) with relation \( r \), and the fused graph-level representation \( \mathbf{\mathbf{Z}_G^f} \).

% After calculating the graph-level representation \( \mathbf{z}_{G^r} \) for each subgraph \( G^r \), we fuse the graph-level representations from all relations to obtain a unified graph-level representation \( \mathbf{z}_G \). This fusion process involves concatenating the graph-level representations from all relations and passing them through a multi-layer perceptron (MLP) for dimensionality reduction:

% \[
% \mathbf{z}_G = \text{MLP} \left( \bigoplus_{r \in \mathcal{R}} \mathbf{z}_{G^r} \right),
% \]

% where \( \bigoplus \) denotes concatenation. The MLP is used to further fuse the information and reduce the dimensionality to produce a compact and information-rich graph-level representation, suitable for downstream tasks such as clustering or other graph analysis tasks.

The similarity-aware pooling mechanism identifies node importance, captures structural similarities, and reflects their relationship to the overall graph, refining graph-level representations and improving clustering performance.

\subsection{Graph-level Graph Construction}
To establish meaningful relationships between graph-level representations, we use dynamic graph kernel methods to compute pairwise similarities between graphs. These methods integrate structural features (e.g., node connectivity and subgraph patterns) and attribute information (e.g., node and edge features) to assess the similarity between two graphs. We calculate a similarity score by graph kernel function for each graph pair, which forms the basis for establishing edges in the dynamic graph. The resulting edges reflect the strength of relationships between graphs, with higher similarity scores indicating stronger connections. This process enables the construction of a dynamic graph that captures both local and global structural dependencies, as well as attribute-based relationships. This refined graph structure is essential for downstream tasks like clustering or representation learning, ensuring accurate modeling and preservation of inter-dependencies between graphs. Specifically, given two graph $G_1$ and $G_2$ with structures \( \mathbf{A}^{r_1}\) and \( \mathbf{A}^{r_2} \) from different relations \( r_1 \) and \( r_2 \). Their graph-level representations are $\mathbf{Z}_{G}^{r_1}$ and $\mathbf{Z}_{G}^{r_2}$. Then, we compute their similarity using a graph kernel function, which evaluates both node features and structural patterns as
\begin{equation}
K(G_1, G_2) = \sum_{v_1 \in G_1, v_2 \in G_2} \phi(\mathbf{z}^{r_1}(v_1)) \cdot \phi(\mathbf{z}^{r_2}(v_2)),
\end{equation}
where \( \phi \) is a feature mapping function that transforms graph representations into a suitable feature space. This kernel function $K(\cdot)$ compares the graph-level features by Eq. (7). The result is a similarity between the two graph representations, which can be interpreted as the strength of their relationship.

\textbf{Dynamic Graph Kernel for Multiple Relations}
In the case of multi-relational graphs, we apply the graph kernel to the set of graph-level representations \( \mathbf{z}_G^{r} \) for each relation \( r \in \mathcal{R} \). The graph kernel computation is extended to capture the interactions between multiple relations:
\begin{equation}
K_\phi(G_1, G_2) = \sum_{r_1, r_2 \in \mathcal{R}} K(G_1^{r_1}, G_2^{r_2}),
\end{equation}

where \( K_\phi(G_1^{r_1}) \) is the graph kernel between the graph-level representations of the graphs \( G_1 \) for relations \( r_1 \) and \( r_2 \), respectively. This formulation ensures that the similarity between two graphs is not only determined by their features but also by their relationships across different relations.

\textbf{Edge Construction Based on Similarity}  
Using graph similarities from the dynamic graph kernel, we construct edges between graphs. For each pair \( G_1 \) and \( G_2 \), an edge is defined as \( \mathbf{\widetilde{A}}_{G_1, G_2} = K(G_1, G_2) \), ensuring connections only between highly similar graphs based on both structural and feature similarities. This approach effectively captures graph relationships by leveraging dynamic graph kernels to construct edges. It preserves both node-graph similarity and structural dependencies, providing a richer and more comprehensive representation of the data.

% \textbf{Constructing the Dynamic Graph-level Representation} The final step is to construct a dynamic graph \( G_{\text{dynamic}} = (V_{\text{dynamic}}, E_{\text{dynamic}}) \), where \( V_{\text{dynamic}} \) represents the set of graphs (nodes) and \( E_{\text{dynamic}} \) represents the set of edges formed based on the computed similarities. Each node in \( V_{\text{dynamic}} \) corresponds to a graph-level representation \( \mathbf{z}_G \), and an edge \( (G_1, G_2) \) exists in \( E_{\text{dynamic}} \) if their similarity exceeds the threshold \( \delta \).

\subsection{Optimization Objective}
To guide clustering and optimize graph-level representations, we propose three loss functions to guide clustering and optimize graph-level representations.

\textbf{Cluster Loss.} The main goal of this loss is to refine the graph-level representations, ensuring they are placed closer to their corresponding cluster centers in the embedding space. The loss is expressed as:
\begin{equation}
\mathcal{L}_{\text{clu}} = - \sum_{i=1}^k \sum_{j=1}^n y_{ij} \log \left( \frac{\exp((\mathbf{z}')^\top \mathbf{c}_i)}{\sum_{l=1}^k \exp((\mathbf{z}')^\top \mathbf{c}_l)} \right),
\end{equation}

where \( \mathbf{z}_G' \) represents the graph-level representation, \( \mathbf{c}_i \) is the centroid of the \( i \)-th cluster, and \( y_{ij} \) is an indicator variable that is 1 if graph \( j \) belongs to the \( i \)-th cluster and 0 otherwise. This loss function encourages the model to pull graph representations closer to their true cluster centers.

\textbf{Adaptive Contrastive Loss} 
After constructing the graph-level graph, we refine these relationships by the contrastive loss. We propose an adaptive contrastive loss, which refines the graph-level relationships by treating edge weights as continuous similarity measures. The loss is expressed as
\begin{equation}
\mathcal{L}_{\text{con}} = \sum_{i=1}^n \sum_{j=1}^n \mathbf{\widetilde{A}}_{G_i,G_j} \, \ell(\mathbf{Z}^i, \mathbf{Z}^j),
\end{equation}
where, the \( \ell \) is the MSE loss, $\mathbf{Z}^i$ and $\mathbf{Z}^j$ denote different graph-level representations with relation $i$ and $j$, respectively.

\textbf{Graph Similarity Loss} To capture the latent consensus structure across multiple relationships, we introduce a graph similarity loss based on a kernel function. The loss is expressed as
\begin{equation}
\mathcal{L}_{\text{sim}} = \sum_{u,v \in V} \left( \text{K}(G_u, G_v) - \mathbf{A}_{uv} \right)^2,
\end{equation}
where \( \text{K}(G_u, G_v) \) is the similarity between the graphs \( G_u \) and \( G_v \) computed by the graph kernel method, and \( \mathbf{A}_{uv} \) is the corresponding edge weight in the dynamic graph. Finally, the total loss is a combination of the three, as $
\mathcal{L} =  \mathcal{L}_{\text{clu}} + \lambda \mathcal{L}_{\text{con}} + \mu \mathcal{L}_{\text{sim}},
$ where \( \lambda \) and \( \mu \) are hyperparameters that control the importance of each loss term. $\mathcal{L}_{\text{clu}}$ encourages graph representations to be close to their corresponding cluster centers, promoting better clustering performance. $\mathcal{L}_{\text{con}}$ ensures that similar graphs are positioned close together in the embedding space, refining the representations. $\mathcal{L}_{\text{sim}}$ aligns the learned graph representations with the graph similarities computed using graph kernels, ensuring meaningful representation learning. In consequence, these losses preserve graph similarity and improve clustering performance.

\begin{table}[ht]
\centering
\caption{Statistics of datasets.}
\label{tab:dataset_statistics}
\begin{tabular}{@{}lcccccc@{}}
\toprule
\textbf{Datasets}        & \textbf{Category}    & \textbf{\#Class} & \textbf{\#Graph} & \textbf{\#Node} & \textbf{\#Edge} \\ \midrule
COX2      & Molecules       & 2  & 467    & 41.22    & 43.45    \\
BZR       & Molecules       & 2  & 405    & 35.75    & 38.36    \\
IMDB-B    & Molecules       & 2  & 188    & 17.93    & 39.45    \\ \midrule
COLLAB   & Social Networks  & 3  & 5000   & 74.49    & 2457.78  \\\midrule
Letter   & Computer Vision  & 15  & 2250   & 4.68     & 3.13     \\ \bottomrule
\end{tabular}
\end{table}

\begin{table*}[!ht]
\centering
\caption{Clustering performance of seven methods on five datasets (means ± std). The red and blue values represent the best and second-best results, respectively.}\small
\renewcommand{\arraystretch}{1} % Adjust row spacing
\setlength{\tabcolsep}{0.9mm}{
\begin{tabular}{c|c|ccc|ccc|ccccc}
\hline
\hline
Metric & Dataset & SP & GK & WL-OA& InfoGraph & GraphCL & JOAO & GFW+KM & GFW+SC & GLCC & DGLC \footnote{1} & Ours \\
\hline
\multirow{5}{*}{ACC}    & BZR   & 78.5±0.0 & 53.9±1.2 &69.6±0.0 & 63.6±2.4  & 71.4±4.1  & 72.6±4.2 & 53.0±0.3 & 52.7±0.8 & 63.6±9.8 & \textcolor{blue}{80.9±0.2} & \textcolor{red}{82.1±1.9} \\
& COX2  & 52.0±0.0 & 50.2±0.0 & 50.8±0.0 & 56.7±0.0  & 68.9±0.5  & 76.5±0.2 & 57.6±4.1 & 58.8±4.5 & 77.3±1.1 & \textcolor{blue}{78.2±0.2} & \textcolor{red}{80.2±0.2} \\
& IMDB  & 53.9±1.1 & 50.2±0.0 & 50.9±0.5 & 50.2±3.0  & 54.6±0.6  & 50.9±2.7 &56.9±2.6 & 51.9±3.6 & \textcolor{blue}{66.6±2.6} & 66.5±0.5 & \textcolor{red}{69.0±1.6} \\
& COLLAB& 38.4±0.1 & 37.5±0.0 & 32.4±1.3 & 37.5±0.0  & 43.3±0.2  & 43.7±5.9 &42.1±0.7 &43.2±0.6 & 55.4±4.0 & \textcolor{blue}{56.8±0.2} & \textcolor{red}{58.4±0.9} \\
& Letter& 15.7±0.4 & 13.4±0.3 & 16.8±0.0 & 13.4±0.3  & 28.4±0.3  & 29.5±5.1  & 19.2±2.4  & 21.5±2.7& 30.1±8.3 & \textcolor{blue}{39.4±0.2} & \textcolor{red}{42.1±0.6} \\
\hline
\multirow{5}{*}{NMI}    & BZR   & 4.1±0.0  & 1.1±1.2 & 5.6±0.0 & 1.6±1.1  & 1.0±0.8  & 1.4±1.1 & 3.4±0.4& 3.4±1.2& 1.2±0.6 & \textcolor{blue}{9.6±1.0}  & \textcolor{red}{9.7±0.2} \\
& COX2  & 0.1±0.4  & 0.0±0.1& 0.5±0.0& 3.3±0.9  & 1.0±0.3  & 1.1±0.3 & 1.5±0.1 & 1.1±0.5 & 1.5±1.3 & \textcolor{blue}{4.4±1.0}  & \textcolor{red}{6.5±0.1} \\
& IMDB  & 6.5±0.0  & 0.3±0.0  & 0.5±0.9& 0.3±0.0  & 5.1±0.3  & 0.3±0.0 
 &1.6±1.0 &0.6±0.8 & \textcolor{blue}{8.1±0.0} & 7.2±0.2  & \textcolor{red}{12.5±0.7} \\
& COLLAB& 2.6±0.1  & 2.5±0.2 & 1.6±0.1 & 3.1±0.0  & 4.2±0.3  & 4.8±4.3 & 3.7±0.3 & 3.5±0.4 & \textcolor{blue}{5.6±1.2} & 5.4±0.3  & \textcolor{red}{6.5±1.1} \\
& Letter& 22.5±0.5 & 15.4±1.0 & 20.9±0.4  & 15.4±1.0  & 30.9±0.4  & 30.6±5.1 & 37.5±1.7 & 38.2±2.4 & 42.6±5.9 & \textcolor{blue}{48.9±0.1} & \textcolor{red}{50.5±0.2} \\
\hline
\multirow{5}{*}{ARI}    & BZR   & 3.9±0.0  & 3.1±3.8 & 0.0±0.0 & 3.6±2.4  & 2.3±1.0  & 3.0±3.3 &0.0±0.0 & 0.0±0.0  & 1.2±0.8 & \textcolor{blue}{6.4±1.2}  & \textcolor{red}{6.9±0.9} \\
& COX2  & 0.1±0.0  & 0.0±0.0 & 0.0±0.0 & 3.3±0.6  & 1.1±0.6  & 1.3±0.0 & 2.1±1.4 & 1.4±1.2 & 0.0±0.0 & \textcolor{blue}{6.8±0.3}  & \textcolor{red}{7.7±0.3} \\
& IMDB  & 0.3±0.0  & 0.7±0.2 & 0.0±0.0 & 4.7±0.0  & 5.1±0.8  & 2.4±3.1  &2.1±1.3 & 0.6±1.2& 8.1±5.2 & \textcolor{blue}{10.5±0.0} & \textcolor{red}{11.4±1.2} \\
& COLLAB& 5.8±0.0  & 2.0±0.0 & 0.9±1.2 & 3.5±0.0  & 4.5±5.9  & 4.5±5.9 & 1.7±1.4  & 2.6±0.5& 6.1±4.0 & \textcolor{blue}{10.8±0.5} & \textcolor{red}{13.1±0.8} \\
& Letter& 12.0±0.0 & 11.7±0.5 & 5.4±1.6 & 13.4±0.3  & 15.4±0.3  & 18.6±2.1   & 17.2±1.9   & 14.1±3.7 & \textcolor{blue}{20.2±2.4} & 17.3±0.2 & \textcolor{red}{24.2±0.1} \\
\hline
\multirow{5}{*}{F1} & BZR   & 66.2±1.4 & 51.4±2.2 & 48.8±2.3  & 60.7±1.2  & 67.5±1.8  & 68.4±0.8 & 47.3±0.2 & 48.6±0.1  & 58.1±0.0 & \textcolor{blue}{76.2±1.0} & \textcolor{red}{80.8±0.5} \\
& COX2  & 46.4±0.4 & 45.3±0.1 & 44.4±4.1  & 53.2±0.4  & 65.3±0.8 & 55.6±1.3 & 51.9±0.4  & 65.9±0.4  & 43.3±1.4 & \textcolor{blue}{74.3±1.1} & \textcolor{red}{79.6±0.8} \\
& IMDB  & 39.2±0.0 & 33.5±0.3  & 36.3±0.1 & 38.7±0.9  & 51.5±0.3  & 48.0±7.3  & 50.1±1.4 & 48.2±1.9 & 63.7±2.3 & \textcolor{blue}{65.8±0.4} & \textcolor{red}{68.7±2.1} \\
& COLLAB& 22.7±0.1 & 24.5±0.4 & 25.5±1.8 & 27.5±0.7  & 38.6±0.5  & 42.5±4.3  & 38.7±0.3 & 39.2±1.4 & 43.4±1.7 & \textcolor{blue}{51.2±0.2} & \textcolor{red}{56.7±0.4} \\
& Letter& 14.5±1.8 & 11.4±0.7 & 13.6±3.1  & 11.3±0.2  & 24.2±2.6  & 26.5±0.5 & 15.4±1.8 & 17.6±1.6 & 25.2±4.2 & \textcolor{blue}{40.2±1.5} & \textcolor{red}{41.8±0.6} \\
\hline
\hline
\end{tabular}}
\label{tab:compare}
\end{table*}

\section{Experiment}

\subsection{Comparison Experiments}
\subsubsection{Baselines}
To evaluate the effectiveness of our proposed method, we compare MGSN with several state-of-the-art baselines, which can be grouped into three main categories: 
(i) Graph Kernel: SP \cite{Borgwardt2005}, GK \cite{Shervashidze2009}, WL-OA, (ii) Unsupervised Graph Representation Learning: InfoGraph \cite{sun2019infograph}, GraphCL \cite{You2020}, JOAO \cite{you2021graph}, (iii) Graph-Level Clustering: GWF \cite{Xu2022}, UDGC, GLCC \cite{ju2023glcc}, DGLC \cite{cai2024deep}. Note that the experimental parameters of InfoGraph \cite{sun2019infograph}, GraphCL \cite{You2020}, JOAO \cite{you2020graphcl}, GLCC \cite{ju2023glcc}, UDGC, and DGLC \cite{cai2024deep} are the same as ours to guarantee a fair comparison.

\subsubsection{Experiment Settings}
The experiments are performed on a Windows-based PyTorch platform using an Nvidia GeForce 4070S GPU and Intel i9 CPU. The learning rate is \(1 \times 10^{-3}\), with a three-layer network architecture and the batch size is 128. The model ran for 50 epochs with early stopping. Results are averaged over ten independent runs.

\subsubsection{Benchmark Datasets}
Experiments are conducted on five publicly available datasets from TUDataset: BZR \cite{cai2024dual}, COX2 \cite{sutherland2003spline}, IMDB-B \cite{yanardag2015deep}, COLLAB \cite{cai2022efficient}, and Letter-low \cite{cai2021graphclustering}, with detailed statistics in Table \ref{tab:dataset_statistics}.

\begin{table*}[h!]
\caption{Sub-relation ablation of the multi-relation Graph: evaluating the performance impact of using a single relation in isolation.}
\centering
\begin{tabular}{c|c|c|c|c|c|c|c|c|c|c}
\hline
\multicolumn{3}{c}{Relations} & \multicolumn{4}{|c|}{\textbf{BZR}} & \multicolumn{4}{|c}{\textbf{COX2}}  \\ \hline
 $\phi$& r & f & ACC & NMI & ARI & F1 & ACC & NMI & ARI & F1  \\ \hline
 \checkmark & -  & -  & 
76.6 $\pm$ 2.1  &  6.3 $\pm$ 0.2 & 5.4 $\pm$ 1.1 &  68.2 $\pm$ 1.6 &
76.4 $\pm$ 1.7  &  3.7 $\pm$ 0.2 & 6.3 $\pm$ 0.2 & 67.9$\pm$ 1.4\\ \hline

-  & \checkmark &  - & 
78.9 $\pm$ 1.7  &  5.2 $\pm$ 0.1 & 3.7 $\pm$ 0.1 & 68.5 $\pm$ 1.0 &
76.5 $\pm$ 1.5  &  3.1 $\pm$ 0.2 & 6.4 $\pm$ 0.1 & 65.4 $\pm$ 1.3 \\ \hline
- & -  & \checkmark & 
79.2 $\pm$ 1.5  &  7.3 $\pm$ 0.4 & 4.6 $\pm$ 1.6 & 72.4 $\pm$ 1.6 &
75.5 $\pm$ 1.3  &  3.0 $\pm$ 0.1 & 6.2 $\pm$ 0.1 & 63.8
$\pm$ 0.8
\\ \hline
 \checkmark  & \checkmark  & \checkmark & 
82.1 $\pm$ 1.9  &  9.7 $\pm$ 0.2 & 6.9 $\pm$ 0.9 & 80.8 $\pm$ 0.5 &
80.2 $\pm$ 0.2  &  6.5 $\pm$ 0.1 & 7.7 $\pm$ 0.3 & 79.6  $\pm$ 0.5\\ \hline

\multicolumn{3}{c}{Relations} & \multicolumn{4}{|c|}{\textbf{IMDB-BINARY }} & \multicolumn{4}{|c}{\textbf{Letter}}  \\ \hline
 $\phi$& r & f & ACC & NMI & ARI&  F1& ACC & NMI & ARI & F1   \\ \hline
 \checkmark & -  & -  & 
54.4 $\pm$ 1.0  &  6.3 $\pm$ 0.2 & 5.2 $\pm$ 0.0 & 36.2 $\pm$ 1.7 & 
39.6 $\pm$ 0.8  &  40.9 $\pm$ 0.1 &  14.4 $\pm$ 0.1 & 44.1 $\pm$ 0.9\\ \hline

-  & \checkmark &  - & 
60.3 $\pm$ 1.5  &   8.8 $\pm$ 0.7 &  8.4 $\pm$ 0.1 & 50.4 $\pm$ 0.7 &  38.7 $\pm$ 1.0  &  44.2 $\pm$ 0.3 &  13.7 $\pm$ 0.2 & 36.6 $\pm$ 1.5\\ \hline
- & -  & \checkmark & 
66.1 $\pm$ 2.0  &  10.8 $\pm$ 0.2 & 10.7 $\pm$ 0.2 & 61.5 $\pm$ 0.6 &
40.9 $\pm$ 0.9  &  48.6 $\pm$ 0.1 &  22.1 $\pm$ 0.2 &  32.1 $\pm$ 0.4\\ \hline
 \checkmark  & \checkmark  & \checkmark & 
69.0 $\pm$ 1.6  &  12.5 $\pm$ 0.7 & 11.4 $\pm$ 1.2 & 68.7 $\pm$ 2.1& 
41.8 $\pm$ 0.3  &  50.5 $\pm$ 0.2 & 24.2 $\pm$ 0.1  & 41.8 $\pm$ 0.6\\ \hline

\end{tabular}
\label{tab:sub-relation}
\end{table*}
\subsection{Performance Indicators}
We present the clustering results of MGSN on five benchmark datasets. The clustering performance is evaluated using four public metrics: Accuracy (ACC) \cite{CMSCGC, zhao2024parameter}, Normalized Mutual Information (NMI) \cite{tu2021deep}, Adjusted Rand Index (ARI) \cite{TGC_ML_ICLR}, and F1 (F1) \cite{tu2024attribute}. The highest and second highest results are highlighted in \textcolor{red}{red} and \textcolor{blue}{blue}, respectively, and are shown in Table~\ref{tab:compare}. The main findings are summarized as follows:
% \footnotetext[1]{Since the original article does not provide experimental results for all datasets relevant to our study, the presented results are derived from our reproduction of the experiments.}
\begin{itemize}
    % \item MGSN consistently outperforms existing methods across multiple datasets, showing that integrating multi-relation modeling to capture a wider range of graph relationships enhances the quality of learned embeddings. 
    % \item MGSN effectively overcomes the limitation of existing methods in that they cannot capture the underlying complex structure relationships in graphs. By leveraging graph kernel strength, MGSN can perceive subtle latent features, leading to better clustering performance.
    \item MGSN consistently surpasses existing methods across multiple datasets, demonstrating that the integration of multi-relation modeling not only enhances the quality of learned embeddings by capturing a broader spectrum of graph relationships but also effectively addresses the limitations of prior approaches in modeling complex structural dependencies. 
    % MGSN discerns subtle latent features by strength of graph kernels, thereby achieving superior clustering performance.
    
    \item MGSN excels on datasets lacking graph node attributes, as traditional clustering methods depend heavily on node-specific features. This advantage arises from its capability to capture and infer the multi-relation structures inherent in the graph. Unlike deep methods that depend on node attributes for clustering, MGSN focuses on the relational dynamics between nodes across multiple views, allowing it to uncover meaningful patterns and semantic relationships even without node-level information.
    % \item MGSN excels in multi-class datasets, mainly due to its relation-aware representation enhancement strategy, which effectively helps the model distinguish cluster boundaries.
    % \item Compared to spectral clustering, our approach leverages deep learning for better performance, overcoming graph kernel methods' limitations in handling complex structures, scalability, and noise sensitivity.
    \item Compared to deep learning methods based on representation learning, our approach maintains a clear advantage. This is due to the integration of dynamic graph kernels, which enable a more detailed recognition of potential relationships across multiple graphs. The dynamic graph kernel enhances the model's ability to capture intricate and latent structural dependencies, offering a more nuanced representation of graph data than previous methods. By combining the two, our method effectively addresses challenges such as scalability and the accurate identification of complex graph relationships.

      % \item Additionally, the graph kernel mechanism provides a more nuanced understanding of graph similarity, considering both structural and attribute information. This combination enables leads to more expressive and context-aware embeddings, which in turn enhances clustering performance.
    
    % \item By combining multi-relation information with graph kernel techniques, MGSN can capture a diverse set of semantic relationships within the graph data. This integration enriches the representation space, ensuring both consistency and discriminability in the learned embeddings. This shows that our method can extract more meaningful features from the graph structure, which improves the clustering results.
    
    % \item  Our relation-aware representation refinement strategy, which adaptively aligns multi-relation information and progressively fuses complementary features through a kernel-guided alignment mechanism, improves the robustness of the embeddings. This mechanism preserves structural nuances in the graph data.
\end{itemize}

\begin{figure}[htbp]
    \centering
    \includegraphics[width=1\linewidth]{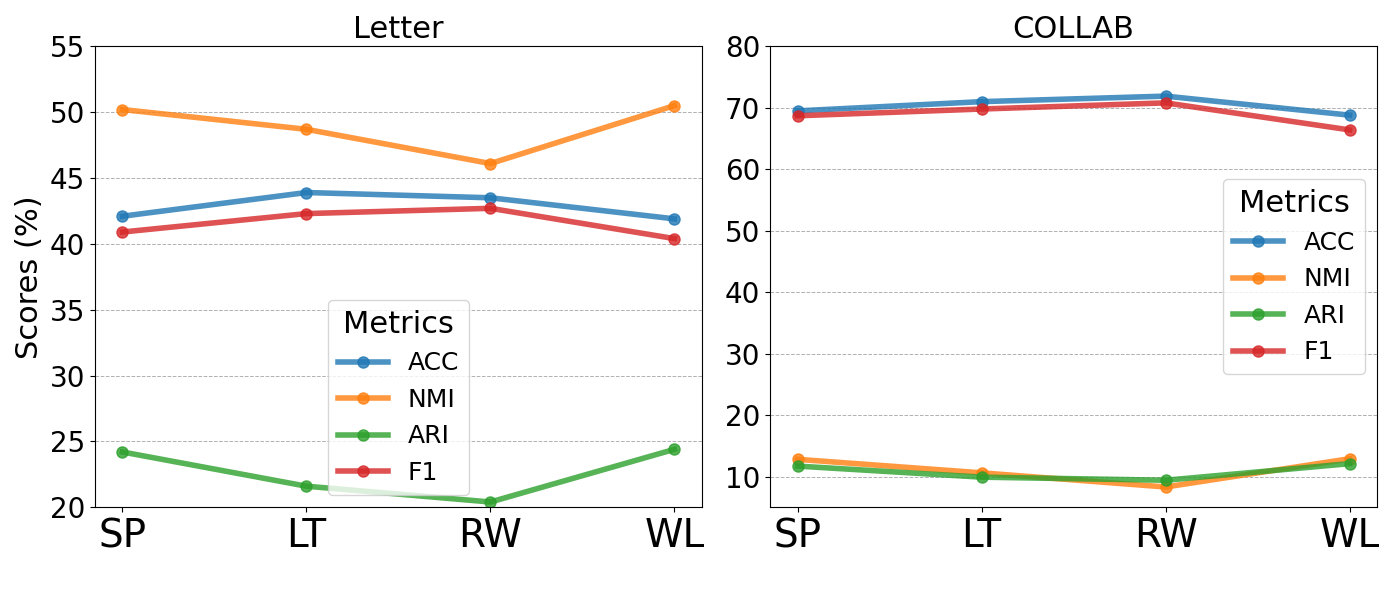}
    \caption{Performance comparison of different graph kernels on Letter dataset.}
    \label{fig:gk}
    \vspace{-10pt}
\end{figure}

\begin{figure*}[htbp]
    \centering
    \includegraphics[width=1\linewidth]{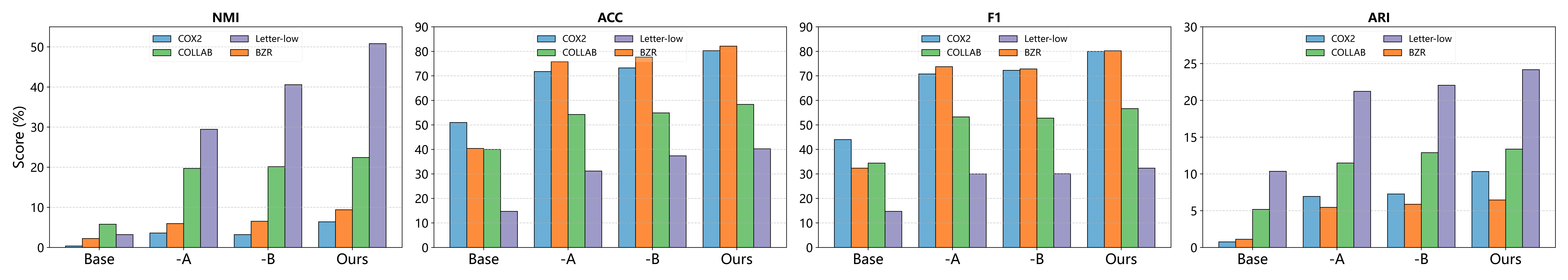}
    \caption{Sub-module ablation experiment results of MGSN, -A and -B represent the use of multi-relation graph and relation-aware representation strength strategy respectively.}
    \label{fig:sub_module}
\end{figure*}

\begin{figure*}[!htbp]
    \centering
    \vspace{-10pt}
    \includegraphics[width=1\linewidth]{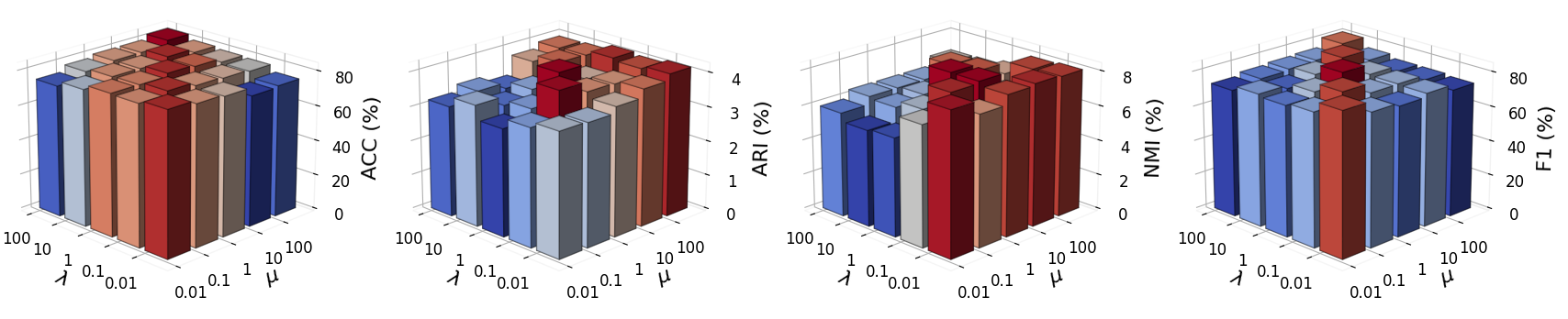}

    \caption{The impact of $\mu$ and $\lambda$ to the performance on COX2. $\mu$ and $\lambda$ changes in the range of [0.01, 100].}
    \label{fig:hyperparam0}
\end{figure*}

\begin{figure*}[!htbp]
    \centering
        \vspace{-10pt}
    \includegraphics[width=1\linewidth]{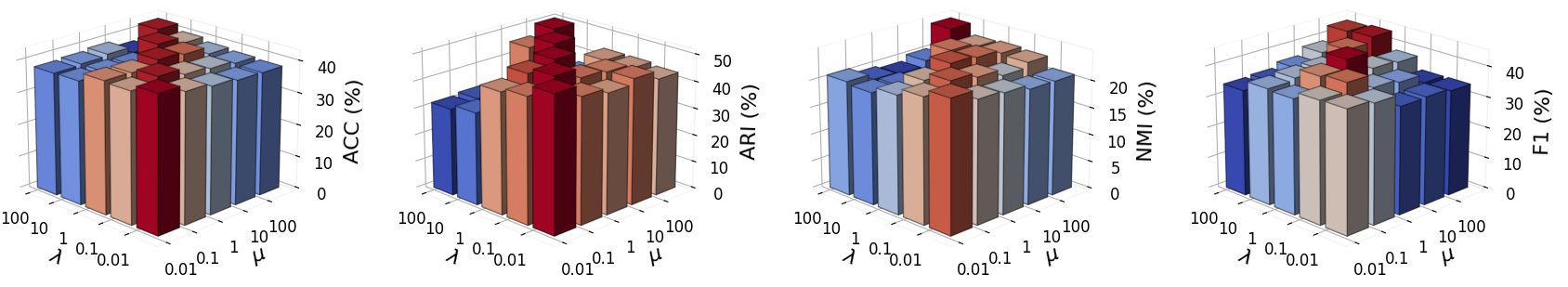}

    \caption{The impact of $\mu$ and $\lambda$ to the performance on Letter-low. $\mu$ and $\lambda$ changes in the range of [0.01, 100].}
    \label{fig:hyperparam1}
\end{figure*}

% \begin{figure*}[!htbp]
%     \centering
%     \begin{subfigure}{0.5\linewidth}
%         \centering
%         \includegraphics[width=1\linewidth]{./images/para_cox2_1.PNG}
%         \label{fig:hyperparam0}
%     \end{subfigure}%
%     \begin{subfigure}{0.5\linewidth}
%         \centering
%         \includegraphics[width=1\linewidth]{./images/para_letter.PNG}
%         \label{fig:hyperparam1}
%     \end{subfigure}
%     \caption{Impact of $\mu$ and $\lambda$ on model performance. Both $\mu$ and $\lambda$ vary in the range of [0.01, 100].}
%     \label{fig:hyperparam}
% \end{figure*}

\subsection{Analysis of Ablation Studies}
\subsubsection{Sub-relation Ablation} To assess the impact of features from different relations on experimental performance. We consider one relation at a time to evaluate. The results are shown in Table~\ref{tab:sub-relation}, and the following conclusions are obtained:

\begin{itemize}
    \item Multi-relation modeling harnesses the strengths of different relations, yielding a more comprehensive graph-level representation. Retaining only one relation weakens clustering performance, notably in NMI and ARI, as the absence of multi-relation consensus fails to effectively reduce intra-cluster distances.  
    \item The fusion relation outperforms individual relations, demonstrating its ability to integrate their advantages. Early fusion enhances the model's capacity to capture diverse graph features, benefiting clustering tasks and boosting overall performance.  
    \item Newly constructed relations do not always surpass original ones due to potential noise. However, the subsequent enhancement strategy helps filter out noise, ensuring the model's robustness and improving overall performance.
\end{itemize}

\subsubsection{Module Ablation}  
To evaluate the effectiveness of the proposed module, we conducted an ablation study, as shown in Fig. \ref{fig:sub_module}. The results are as follows:
\begin{itemize}
    \item When only a single part exists, the performance decreases significantly compared to the complete module.
    \item Relation-aware representation strategy is of great significance for capturing multi-dimensional features. Removing this module will not effectively aware and align meaningful clustering signals, resulting in homogeneity in representation, and ultimately significantly deteriorating performance on datasets with simple structures.
    \item When only the relation-aware representation strategy is applied, a decline in clustering performance is observed. This reduction arises because an origin relation fails to provide a sufficiently high-quality structure to support aware enhancement. Additionally, the origin relation introduces noise into the representations, which hinders the alignment of the final multi-view representations and ultimately compromises clustering performance.
    \item When only the multi-relation graph is used, the performance decreases to a certain extent. This is mainly because the relation-aware representation strategy provides further refinement and screening for the fused clustering signal. Removing the hierarchical enhancement module reduces the cohesion of the representation and suboptimal clustering performance.
\end{itemize}

\subsubsection{Graph Kernel Ablation} To comprehensively assess the influence of various graph kernel functions on experimental performance, we conducted multiple experiments using the Letter-lows dataset, employing four distinct graph kernel functions, including SP \cite{Shervashidze2009}, LT \cite{johansson2014global}, RW \cite{borgwardt2005shortest} and WL \cite{shervashidze2011weisfeiler}. The corresponding results are presented in Fig.~\ref{fig:gk}, from which the following conclusions are drawn:
\begin{itemize}
    \item The choice of graph kernel affects the performance of the graph clustering task to some extent. This indicates that the ability of the kernel to capture structural similarities plays a crucial role in improving performance.
    \item The four evaluation metrics may show inconsistent trends with different graph kernels due to their distinct recognition mechanisms. For example, the SP kernel emphasizes subgraph patterns, while the LT method also captures long-range dependencies. Therefore, selecting graph kernels with superior performance is a common practice.
    \item While different kernel choices cause slight performance variations, MGSN maintains consistent results. This stems from our powerful network that effectively adapts to various graph kernels to generate high-quality representations without dependence on any specific kernel.
\end{itemize}

\subsubsection{Hyperparameter Analysis}
To explore the impact of varying loss ratios on clustering performance, we adjusted the hyperparameters \(\mu\) and \(\lambda\) within the range [0.01, 100], using the COX2 and Letter-low datasets for validation. The results, shown in Fig. \ref{fig:hyperparam0} and \ref{fig:hyperparam1}, reveal that a balanced loss ratio of \(\mu\) to \(\lambda\) at 1:1 achieves optimal performance.

\section{Conclusions}
In this paper, we introduce a novel clustering network termed MGSN to address the challenges in graph-level clustering. The proposed framework integrates multi-relation modeling with graph kernel techniques, allowing the model to capture diverse semantic relationships between nodes and graphs. By applying a relation-aware representation refinement strategy and kernel-guided feature alignment, MGSN improves graph-level representations and enhances clustering performance. Extensive experiments on multiple benchmark datasets show that MGSN outperforms state-of-the-art methods, demonstrating its ability to effectively combine multi-relation structures and graph kernel features. Additionally, this integration enhances the consistency and discriminability of graph-level embeddings, making it more robust to complex, multi-relational data. Future work will focus on optimizing the model’s scalability and extending its application to larger, more complex graph-level clustering tasks.

\bibliographystyle{IEEEtran}
\bibliography{refs, ndut}
\end{document}